%% file: main.tex
\documentclass[journal]{IEEEtran}
\usepackage{amsmath,amsfonts}
\usepackage{algorithmic}
\usepackage{algorithm}
\usepackage{array}
\usepackage[caption=false,font=normalsize,labelfont=sf,textfont=sf]{subfig}
\usepackage{textcomp}
\usepackage{stfloats}
\usepackage{url}
\usepackage{verbatim}
\usepackage{graphicx}
\usepackage{cite}

\input{preamble}

\begin{document}

\title{
Finding Optimal Video Moment without Training: Gaussian Boundary Optimization for Weakly Supervised Video Grounding
}


\author{Sunoh Kim, Kimin Yun, Daeho Um
\thanks{The present research was supported by the research fund of Dankook University in 2025. (corresponding author: Daeho Um.)}
\thanks{S. Kim is with the Department of Computer Engineering, Dankook University, Yongin, 
South Korea (e-mail: suno8386@dankook.ac.kr)}
\thanks{K. Yun is with Electronics and Telecommunications Research Institute and the University of Science and Technology~(UST), Daejeon,
South Korea (e-mail: kimin.yun@etri.re.kr)}
\thanks{D. Um is with the School of Electrical and Computer Engineering, University of Seoul, Seoul, South Korea (e-mail: daehoum@uos.ac.kr).}
}

\markboth{IEEE TRANSACTIONS ON MULTIMEDIA,~2026}%
{Kim \MakeLowercase{\textit{et al.}}: Finding Optimal Video Moment without Training: Gaussian Boundary Optimization for Weakly Supervised Video Grounding}



\maketitle

\begin{abstract}
\input{0-abstract}

\end{abstract}


\section{Introduction}
\input{1-intro}

\section{Related Work}
\input{2-related-work}

\section{Proposed Method}
\input{3-method}

\section{Experiment}
\label{sec:experiment}
\input{4-experiment}

\section{Conclusion}
\input{5-conclusion}

\section*{Acknowledgments}
\input{6-acknowledgements}

\input{7-appendix}

\bibliographystyle{IEEEtran}
\bibliography{references}

\vfill

\end{document}

%% file: preamble.tex
\usepackage[dvipsnames]{xcolor}

\usepackage{newfloat}
\usepackage{listings}

\usepackage{amssymb}
\usepackage{booktabs}
\usepackage{soul}
\usepackage{multirow}
\usepackage{hyperref}

\usepackage{amsthm}
\newtheorem{theorem}{Theorem}

\theoremstyle{definition}

\theoremstyle{remark}

\usepackage{xcolor}
\definecolor{softblue}{RGB}{65,105,225}

\usepackage{pifont}

\renewcommand{\paragraph}[1]{
     \noindent{\textbf{#1}} 
 }
 
\usepackage[cjk]{kotex}

\usepackage[capitalize]{cleveref}
\crefname{section}{Sec.}{Secs.}
\Crefname{section}{Section}{Sections}
\Crefname{table}{Table}{Tables}
\crefname{table}{Tab.}{Tabs.}

\newcommand{\etal}{\textit{et al}. }
\newcommand{\ie}{\textit{i}.\textit{e}., }
\newcommand{\eg}{\textit{e}.\textit{g}., }

\usepackage{bm}

%% file: 0-abstract.tex
Weakly supervised temporal video grounding aims to localize query-relevant segments in untrimmed videos using only video-sentence pairs, without requiring ground-truth segment annotations that specify exact temporal boundaries. 
Recent approaches tackle this task by utilizing Gaussian-based temporal proposals to represent query-relevant segments.
However, their inference strategies rely on heuristic mappings from Gaussian parameters to segment boundaries, resulting in suboptimal localization performance. To address this issue, we propose Gaussian Boundary Optimization (GBO), a novel inference framework that predicts segment boundaries by solving a principled optimization problem that balances proposal coverage and segment compactness. 
We derive a closed-form solution for this problem and rigorously analyze the optimality conditions under varying penalty regimes. Beyond its theoretical foundations, GBO offers several practical advantages: it is training-free and compatible with both single-Gaussian and mixture-based proposal architectures. 
Our experiments show that GBO significantly improves localization, achieving state-of-the-art results across standard benchmarks. Extensive experiments demonstrate the efficiency and generalizability of GBO across various proposal schemes.
The code is available at \href{https://github.com/sunoh-kim/gbo}{https://github.com/sunoh-kim/gbo}.

%% file: 1-intro.tex

Video grounding is an important yet challenging task that requires a machine to watch a video and localize the temporal segment corresponding to a given query~\cite{mun2020local,zeng2020dense,kim2021plrn, yuan2019find, kim2022swag}. 
In the weakly supervised setting, where ground-truth boundary annotations are unavailable during training, recent approaches have developed various proposal representations to bridge the semantic gap between video and language modalities~\cite{huang2021cross, tan2021logan, mithun2019weakly, wang2021weakly, kim2024learnable,zhang2020counterfactual,kong2023dynamic, lv2023counterfactual, yoon2023scanet, bao2024omnipotent, kim2025enhancing, ma2023dual, lv2025variational, tang2025dual, li2025etc, zheng2022cnm, zheng2022cpl, kim2024gaussian}. Among these approaches, Gaussian-based temporal proposals~\cite{kong2023dynamic, lv2023counterfactual, yoon2023scanet, bao2024omnipotent, kim2025enhancing, ma2023dual, lv2025variational, tang2025dual, li2025etc, zheng2022cnm, zheng2022cpl, kim2024gaussian} have demonstrated superior performance in modeling temporal structures by offering a flexible representation.

Early studies~\cite{zheng2022cnm, zheng2022cpl} introduce single-Gaussian proposals learned during training, which significantly reduce the number of candidate segments compared to previous non-Gaussian methods. Despite the potential, these single-Gaussian proposals are inherently unimodal and symmetric, limiting their effectiveness in modeling diverse and asymmetric temporal patterns.
To address this limitation, Gaussian mixture proposals~\cite{kim2024gaussian} have been introduced, combining multiple Gaussians to capture a broader range of temporal events.
However, while existing Gaussian-based methods focus heavily on modeling temporal events and optimizing their networks during training, they largely overlook the inference stage.
Specifically, when converting a Gaussian proposal into a temporal segment during inference, these methods typically adopt a simple heuristic: using the Gaussian mean as the segment center and defining the segment duration based on the Gaussian standard deviation (\ie Gaussian width).
This predefined mapping fails to fully leverage the rich structural information encoded in the proposal and cannot guarantee optimal results.
Consequently, such inference strategies often lead to suboptimal localization performance.


\begin{figure}[t!]
  \centering
  \includegraphics[width=\linewidth]{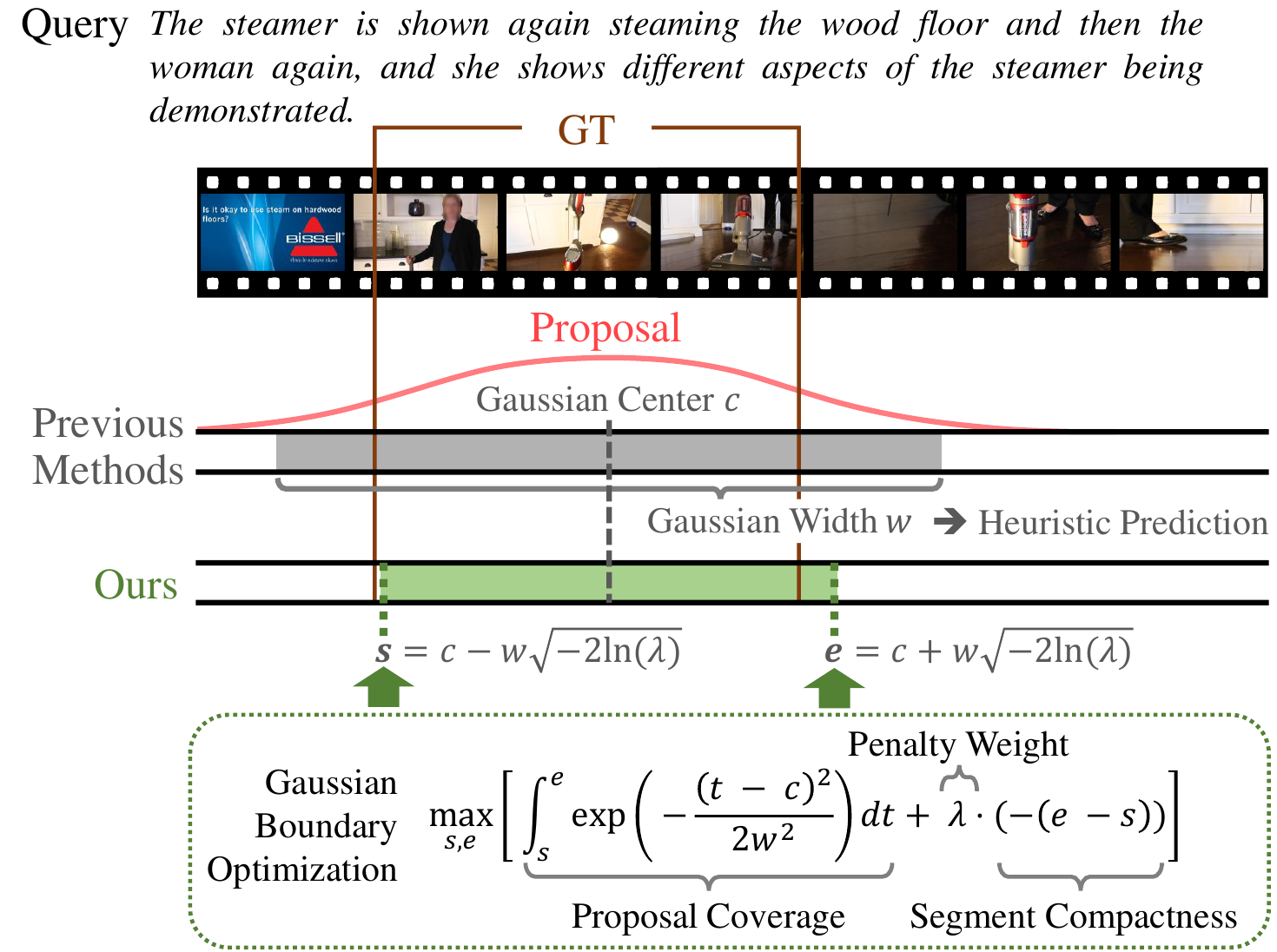}
  \caption{Weakly supervised video grounding. Compared to previous methods based on heuristic segment prediction, our Gaussian boundary optimization formulates the segment prediction as a principled optimization problem. By maximizing coverage under a Gaussian-based proposal while penalizing excessive segment length via a penalty weight, our method yields segments that more accurately align with the query-relevant content.
  }
\label{fig-concept-art}
\end{figure}

To overcome these limitations, we propose Gaussian Boundary Optimization (GBO), a novel inference technique that predicts segment boundaries through principled optimization. 
For this optimization, we define two key concepts: \textit{proposal coverage} and \textit{segment compactness}. 
\textit{Proposal coverage} is defined as the area under the Gaussian proposal curve within the selected segment, quantifying how well the predicted segment aligns with the underlying proposal.
\textit{Segment compactness} refers to the temporal tightness of the segment, measured by its length, which penalizes overly long intervals that might include irrelevant content.
Building on these concepts, GBO formulates boundary prediction as the maximization of an objective function that explicitly balances proposal coverage and segment compactness, as illustrated in \cref{fig-concept-art}. This trade-off is controlled by a penalty weight $\lambda \ge 0$, where smaller $\lambda$ values encourage broader coverage while larger values favor more compact segments.
We provide a comprehensive theoretical analysis of this objective and characterize the optimal solution based on the value of $\lambda$. We demonstrate that when $0 \le \lambda < 1$, the solution yields a non-degenerate segment centered at the peak of the proposal, with closed-form expressions for its boundaries. In contrast, when $\lambda \ge 1$, the solution collapses to a degenerate interval at the proposal center, reflecting the stronger preference for compactness.

\begin{figure}[t!]
  \centering
  \includegraphics[width=\linewidth]{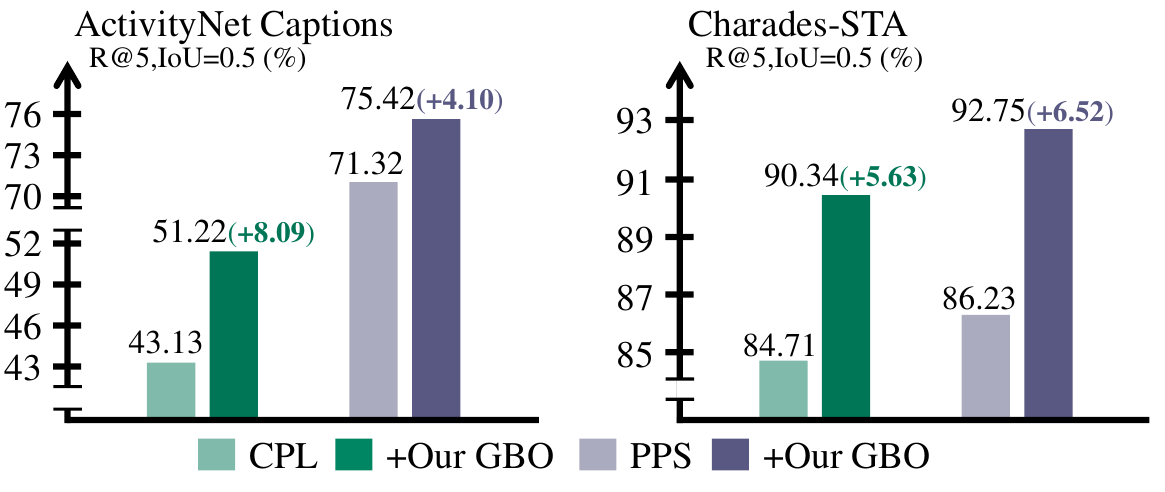}
  \caption{
  Performance improvement by applying our Gaussian Boundary Optimization to existing Gaussian proposal-based methods. Our method yields significant gains over both CPL and PPS baselines, improving performance by up to +8.09\%p on ActivityNet Captions and +6.52\%p on Charades-STA, respectively.
  }
\label{fig:performance-increase}
\end{figure}

A key strength of GBO lies in its model-agnostic design. It requires no modifications to the training procedure and can be seamlessly integrated into a wide range of existing Gaussian proposal-based frameworks.
Importantly, GBO is compatible with both single-Gaussian proposals (\ie CPL~\cite{zheng2022cpl}) and Gaussian mixture proposals (\ie PPS~\cite{kim2024gaussian}), enabling it to enhance more expressive proposal representations, as demonstrated in ~\cref{fig:performance-increase}.
We validate the generality and effectiveness of GBO through comprehensive experiments on three representative Gaussian proposal-based architectures~\cite{zheng2022cnm,zheng2022cpl,kim2024gaussian} across widely-used benchmark datasets: Charades-STA~\cite{gao2017tall} and ActivityNet Captions~\cite{krishna2017dense}. In all cases, GBO consistently improves localization performance and achieves state-of-the-art results.
This improvement is achieved without any additional training and with negligible inference overhead, making GBO a lightweight yet powerful enhancement.
Furthermore, we conduct an extensive empirical analysis of the penalty weight, revealing its critical role in balancing coverage and compactness for optimal segment prediction.

Overall, GBO provides a theoretically sound and practically effective framework for proposal-based localization in weakly supervised video grounding.
Our main contributions are summarized as follows:

\begin{itemize}
    \item We propose Gaussian Boundary Optimization (GBO), a novel inference framework for weakly supervised video grounding that formulates segment prediction as a principled optimization problem balancing proposal coverage and segment compactness, addressing the limitations of existing heuristic inference strategies.


    \item We provide a complete mathematical foundation for GBO, including closed-form solutions to the optimization problem under different penalty weight regimes and rigorous theoretical analysis of optimality conditions. Through comprehensive case analysis, we formally prove the conditions under which the optimal solution yields a non-degenerate segment.


    \item We demonstrate that GBO is a model-agnostic, training-free inference framework that seamlessly integrates with any Gaussian proposal-based method, including both single-Gaussian and Gaussian mixture representations. Our extensive experiments show that GBO consistently improves localization performance across diverse architectures and datasets, yielding significant gains of up to 11.25\%p. GBO-enhanced models achieve state-of-the-art results without additional training and with negligible inference overhead, making GBO a practical and powerful inference framework.

\end{itemize}

%% file: 2-related-work.tex
\subsection{Fully Supervised Video Grounding}
Fully supervised video grounding (FSVG) aims to localize video segments for a natural language query using precise temporal annotations. Progression has moved from sliding-window proposals~\cite{gao2017tall,anne2017localizing} to query-guided and graph-aware proposals that raise candidate quality~\cite{liu2018attentive,zhang2019cross,xu2019multilevel,chen2019semantic,xiao2021boundary,zhang2019man,liu2021context}, with dense temporal pairing broadening coverage~\cite{zhang2020learning}. In parallel, proposal-free models directly regress boundaries with stronger video–text interaction~\cite{ghosh2019excl,zeng2020dense,li2021proposal,zhao2021cascaded,mun2020local,yuan2019find}, and are complemented by boundary-aware prediction, span-based localization, memory-guided semantic learning, cross-modal NAS, dynamic modulation, and early sequence modeling~\cite{wang2020temporally,zhang2020vslnet,liu2022memory,yang2022video,yuan2019semantic,chen2018temporally}.
Language structure and semantic word graphs sharpen alignment~\cite{kim2021plrn,kim2022swag}, while unified and scalable designs improve generalization and efficiency~\cite{lin2023univtg,mu2024snag}. 
Transformer~\cite{vaswani2017attention}-based architectures provide end-to-end set prediction with strong boundary handling~\cite{lei2021detecting,chen2021end,shi2024end,lee2024bam}. 
Recent advances further enhance visual–textual grounding through motion-aware fusion and appearance modeling~\cite{liu2023exploring}, diffusion-based fine-grained boundary reasoning~\cite{liu2023conditional}, and tracking-inspired memory architectures~\cite{xiong2024rethinking}, providing richer temporal dynamics and improved interpretability.
Despite these advances, FSVG still relies on dense temporal labels, which are costly to obtain, prone to boundary noise, and difficult to scale. These limitations have motivated weakly supervised alternatives.

\subsection{Weakly Supervised Video Grounding}
Weakly supervised video grounding (WSVG) aims to localize temporal segments relevant to a given natural language query without requiring explicit temporal annotations during training. Existing methods can be broadly categorized into multiple instance learning (MIL)-based and reconstruction-based approaches.
\textbf{MIL-based approach}~\cite{huang2021cross, zhang2020counterfactual, yang2021local,kim2024learnable,zheng2022cnm, yoon2023scanet} generates semantically irrelevant negative proposals to facilitate the discrimination between positive and negative proposals.
These methods typically employ either inter-video~\cite{zhang2020counterfactual, huang2021cross,yang2021local,Chen_Luo_Zhang_Ma_2022} or intra-video negative~\cite{yoon2023scanet, zheng2022cnm, zheng2022cpl} sampling strategies.
Chen~\etal~\cite{Chen_Luo_Zhang_Ma_2022} generate pseudo labels by exploiting inter-video samples; however, such negative proposals are often insufficiently challenging, as the most confusing segments typically appear within the same video that is semantically aligned with the query. To address the problem, several methods~\cite{yoon2023scanet, zheng2022cnm, zheng2022cpl} construct intra-video negative proposals, introducing harder negatives that are more effective for training.
\textbf{Reconstruction-based approach}~\cite{lin2020weakly, song2020weakly, kong2023dynamic, lv2023counterfactual, yoon2023scanet, bao2024omnipotent, kim2025enhancing, ma2023dual, lv2025variational, tang2025dual, li2025etc, zheng2022cnm, zheng2022cpl, kim2024gaussian} focuses on generating proposals capable of reconstructing masked queries. These techniques operate under the assumption that well-aligned proposals can more accurately reconstruct the original query. They commonly employ sliding windows~\cite{lin2020weakly, song2020weakly} or learnable Gaussian functions~\cite{kong2023dynamic, lv2023counterfactual, yoon2023scanet, bao2024omnipotent, kim2025enhancing, ma2023dual, lv2025variational, tang2025dual, li2025etc, zheng2022cnm, zheng2022cpl, kim2024gaussian} as temporal proposals.
Recent advancements further explore the alignment between visual and textual modalities. Ma~\etal~\cite{ma2023dual} reconstruct masked video clips and queries via two complementary branches, while Tang~\etal~\cite{tang2025dual} disentangles video features into object-aware and motion-aware components, aligning them with nouns and verbs, respectively. To expand pseudo-boundaries, Lin~\etal~\cite{li2025etc} leverage multimodal large language models (MLLMs) for annotations.
Ma~\etal~\cite{ma2022weakly} decouple concept prediction from temporal localization and enforce cross-video consistency for weakly supervised grounding.
Fang~\etal~\cite{fang2024rethinking} reformulate weakly supervised video grounding as a game-theoretic interaction between visual and textual agents, encouraging mutual optimization through competitive collaboration.

\begin{figure*}[t!]
  \centering
  \includegraphics[width=\linewidth]{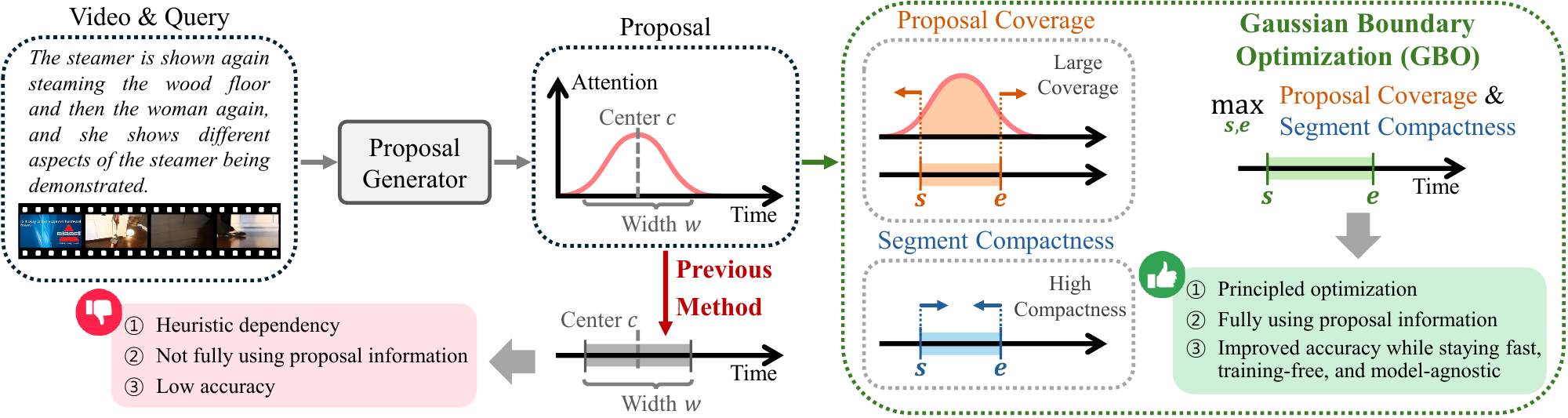}
  \caption{Overview of the proposed Gaussian Boundary Optimization (GBO) framework for video grounding. We first generate Gaussian-based temporal proposals characterized by center and width. Unlike prior heuristic strategies that map proposals to segments with limited proposal information, GBO formulates boundary prediction as an optimization problem balancing proposal coverage and segment compactness. This principled approach fully leverages proposal information, yielding improved accuracy while remaining fast, training-free, and model-agnostic.
  }
\label{fig-framework}
\end{figure*}

\subsection{Gaussian Proposal-Based Methods}
Gaussian-based proposal representations have been widely adopted due to their effectiveness in modeling the temporal structure of video events~\cite{kong2023dynamic, lv2023counterfactual, yoon2023scanet, bao2024omnipotent, kim2025enhancing, ma2023dual, lv2025variational, tang2025dual, li2025etc, zheng2022cnm, zheng2022cpl, kim2024gaussian}.
Early studies~\cite{zheng2022cnm, zheng2022cpl} propose single Gaussian functions as temporal proposals. However, a single Gaussian inherently assumes a unimodal structure, limiting its capacity to capture complex or multi-event moments.
To address this limitation, Gaussian mixture proposals~\cite{kim2024gaussian} have been introduced, enabling more expressive modeling of temporal structures through multiple Gaussian components. Despite these advancements, existing Gaussian proposal-based methods continue to rely on heuristic inference strategies. Specifically, to convert Gaussian proposals into temporal segments, most methods center the segment on the mean of the Gaussian and define its duration based on the Gaussian width. 
While this approach is straightforward, it fails to fully leverage the underlying structure encoded in the Gaussian proposals and cannot guarantee optimal boundary prediction.
Recently, diverse inference strategies have been explored in~\cite{kim2025enhancing}, introducing multiple techniques for boundary prediction and top-1 selection. While this work~\cite{kim2025enhancing} highlights the importance of the inference stage and improves performance, it is only applicable to Gaussian mixture-based methods and still relies on heuristic boundary prediction. In contrast, we formulate boundary prediction as an optimization problem that balances proposal coverage and segment compactness. Our approach is model-agnostic and applicable to both single and mixture Gaussian proposals, improving performance across different architectures and datasets.

%% file: 3-method.tex
\subsection{Motivation and Framework Overview}
\label{sec:overview}
In video grounding, the goal is to precisely predict a query-relevant temporal segment. 
Most prior Gaussian-based approaches~\cite{kong2023dynamic,lv2023counterfactual,yoon2023scanet,bao2024omnipotent,kim2025enhancing,ma2023dual,lv2025variational,tang2025dual,li2025etc,zheng2022cnm,kim2024gaussian,zheng2022cpl} focus on training the model to generate better Gaussian proposals.
For inference, these methods convert the proposals into segments via fixed heuristics, typically centering the segment at the proposal center and setting its length from the proposal width. 
Despite their simplicity, such heuristics fail to exploit the rich structural information in the proposal, resulting in suboptimal segment selection.
We therefore pose the complementary question: \textit{how can we optimally predict the segment boundaries from a given proposal representation?} 
To answer this, we formulate boundary prediction as a principled optimization problem that directly leverages the proposals. The core novelty of our work lies in this shift from \emph{training-time proposal generation} to \emph{inference-time proposal utilization}, which leads to consistent performance improvements while maintaining fast inference.


\cref{fig-framework} illustrates the overall workflow of our method. 
Given an untrimmed video and a textual query, we first generate Gaussian-shaped temporal proposals, parameterized by center $c$ and width $w$, using a transformer-based proposal generator (\cref{sec:preliminary}).
Instead of mapping center and width to a fixed-length segment heuristically, our \emph{Gaussian Boundary Optimization (GBO)} predicts boundaries by maximizing proposal coverage while penalizing segment length (\cref{sec:gbo}).
Under mild conditions, we derive closed-form optimal solutions, yielding the final segment without retraining or architectural changes (\cref{sec:gbo-solution}).
The same principle extends to Gaussian mixture proposals (\cref{sec:mixture}).
Our GBO is training-free, model-agnostic, and computationally efficient with negligible inference overhead, improving localization accuracy across diverse Gaussian-based frameworks (\cref{sec:experiment}).


\subsection{Preliminary}
\label{sec:preliminary}
In weakly supervised video grounding, 
Gaussian proposal-based methods~\cite{kong2023dynamic, lv2023counterfactual, yoon2023scanet, bao2024omnipotent, kim2025enhancing, ma2023dual, lv2025variational, tang2025dual, li2025etc, zheng2022cnm, zheng2022cpl, kim2024gaussian} generate temporal proposals using learnable Gaussian functions, which encode both the center and width of potential events.
We adopt this formulation and generate Gaussian proposals by following strategies introduced in prior methods~\cite{zheng2022cnm,zheng2022cpl,kim2024gaussian}. 
Specifically, we first use pre-trained encoders to extract features from both untrimmed videos and sentence queries.
The video encoder processes videos using 3D convolutional neural networks~\cite{tran2015learning,carreira2017quo}, producing chunk-wise features $\mathbf{V}$. In parallel, the query encoder utilizes GloVe embeddings~\cite{pennington2014glove} to convert sentence queries into word-level features $\mathbf{Q}$.
Given these features $\mathbf{V}$ and $\mathbf{Q}$, we generate candidate proposals using a transformer-based module~\cite{vaswani2017attention}. This module combines $\mathbf{V}$ and $\mathbf{Q}$ by
\begin{math}
  f_{td}(\mathbf{V}, f_{te}(\mathbf{Q}))\text{,}
\end{math}
where the transformer encoder $f_{te}(\cdot)$ processes $\mathbf{Q}$, and the decoder $f_{td}(\cdot)$ takes both $\mathbf{V}$ and $f_{te}(\mathbf{Q})$ as input.
The transformer outputs a sequence of multi-modal representations, with the final token serving as a representative feature.
This representative feature is then used to parameterize the Gaussian proposals by determining the centers $c$ and widths $w$ for each proposal.

\subsection{Gaussian Boundary Optimization: Formulation}
\label{sec:gbo}
This section introduces the objective of GBO, formalizing boundary selection as a trade-off between proposal coverage and segment compactness.

\subsubsection{Statement of the Optimization}
We obtain a Gaussian-shaped proposal from a proposal generator:
\begin{equation}\label{eq:gaussian}
f(t) \;=\; \exp\!\Bigl(-\frac{(t - c)^2}{2\,w^2}\Bigr),
\end{equation}
where $t$ denotes a temporal index, $c$ is the center, and $w>0$ is the width controlling how rapidly $f(t)$ decays as the distance $|t-c|$ increases. Note that $f(t)$ reaches its maximum value of $1$ at $t=c$.

Let $s$ and $e$ denote the start and end boundaries of a video segment, respectively, where $e > s$.
We aim to predict a video segment $[s,e]$ that maximally captures the proposal $f$ while avoiding unnecessarily large segments. To formalize this, we define three key concepts: proposal coverage, segment compactness, and penalty weight.
The \emph{proposal coverage} measures how much of the proposal’s high-confidence mass is retained within the predicted segment, formally defined as $\int_s^e f(t)\,dt$, \ie the area under the curve of the proposal $f$ over $[s, e]$. 
The \emph{segment compactness} reflects the preference for concise intervals, which we encode as the negative segment length $-(e-s)$, so that longer segments are penalized in the objective.
To balance the trade-off between maximizing coverage and minimizing length, we introduce a \emph{penalty weight} $\lambda \ge 0$, which controls the relative importance of segment compactness in our optimization.
We then formulate the objective function:
\begin{equation}\label{eq:objective}
J(s,e) 
\;=\; \underbrace{\int_s^e f(t)\,dt}_{\text{proposal coverage}} \;+\; \lambda \cdot\underbrace{(-(e-s))}_{\substack{\text{segment}\\\text{compactness}}}.
\end{equation}
Our goal is to maximize this objective function:
\begin{equation}\label{eq:max-objective}
\max_{s,e}\; \Bigl[\;\int_{s}^{e} f(t)\,dt \;-\; \lambda\,(e-s)\Bigr],
\end{equation}
where $f(t)$ is given by \eqref{eq:gaussian} and $\lambda \ge 0$.
This optimization formulation explicitly balances the trade-off between segment coverage and temporal compactness.
It encourages the predicted segment to tightly capture the query-relevant content while excluding irrelevant or redundant context.
By linearity of the integral, the objective can be written as
\begin{equation}\label{eq:max-objective-eq}
\max_{s,e}\; \Bigl[\;\int_{s}^{e} \bigl( f(t) \;-\; \lambda\bigr)\,dt\Bigr]
\end{equation}
This form shows that we are maximizing the net area of the \emph{shifted} proposal \(f(t)-\lambda\), \ie the optimal segment is the continuous time span where the proposal stays above the height threshold \(\lambda\).

\subsubsection{Partial Derivatives and Stationary Conditions}
To find a stationary point of the formulated objective function in ~\eqref{eq:max-objective-eq}, we take partial derivatives of $J(s,e)$ with respect to $e$ and $s$. 
This leads to the following necessary condition:
\begin{equation}\label{eq:se-condition}
f(s) \;=\; f(e) \;=\; \lambda.
\end{equation}
Full derivations are provided in the Appendix.
This condition ensures that both ends of the predicted segment lie on the same value of the Gaussian curve.
Recall from \eqref{eq:gaussian} that $f(t)\le 1$ for all $t$, with equality occurring only at $t=c$. Therefore, if $\lambda>1$, the condition $f(t)=\lambda$ cannot be satisfied for any value of $t$. This mathematical constraint indicates a fundamentally different optimal behavior when $\lambda>1$.

\subsubsection{Coverage Bound and Segment Length}
To gain more intuition about our objective function, note that since $0 \le f(t)\le 1$ for all $t$, we can establish:
\[
\int_s^e f(t)\,dt 
\;\le\; (e-s)\cdot \max_{t\in[s,e]} f(t) 
\;\le\; e-s.
\]

Hence,
\begin{equation}
\label{eq:upperbound-objective}
J(s,e) = \int_s^e f(t)\,dt - \lambda (e - s) \le (1 - \lambda)(e - s)
\end{equation}

From this inequality, we can draw important conclusions:
\begin{itemize}
    \item If $\lambda > 1$, then $(1 - \lambda) < 0$, implying that $J(s, e)$ becomes negative for any segment with positive length. In this case, the optimal value of $J$ is achieved by minimizing the segment length, \ie letting $e = s$, which yields $J = 0$.
    \item If $0 \le \lambda \le 1$, the non-negative value of $(1 - \lambda)$ allows for a potentially positive objective value. A suitable segment can therefore be predicted to maximize $J(s, e)$.
\end{itemize}

\subsection{Gaussian Boundary Optimization: Closed-form Solution}
\label{sec:gbo-solution}
We now solve the GBO objective in ~\eqref{eq:max-objective-eq}, deriving closed-form optimal boundaries across the penalty regimes of $\lambda$.

\subsubsection{\texorpdfstring{$\bm{0 \le \lambda < 1}$}{0 <= lambda < 1}}
Since $f(t)$ reaches its maximum value of 1 at $t=c$ and decreases symmetrically in both directions, the equation $f(t)=\lambda$ has exactly two solutions when $0 \le \lambda < 1$. We can derive these solutions as follows:
\begin{align}
\label{eq:solve-lambda}
&\exp\left(-\frac{(t - c)^2}{2w^2}\right) 
= \lambda \nonumber \\
&\Longrightarrow \quad
-\frac{(t - c)^2}{2w^2} 
= \ln(\lambda) \nonumber \\
&\Longrightarrow \quad
(t - c)^2 
= -2w^2 \ln(\lambda).
\end{align}
Since $\ln(\lambda) < 0$ for $0 < \lambda < 1$, we obtain real-valued solutions for $t$:
\begin{equation}
t = c \pm w\,\sqrt{-\,2\,\ln(\lambda)}.
\end{equation}
Let us denote
$
d \;=\; w\,\sqrt{-\,2\,\ln(\lambda)} \;>\; 0.
$
Consequently, the condition in \eqref{eq:se-condition} is satisfied by the symmetric interval:
\begin{equation} \label{eq:optimal-boundary}
s^* = c - d,\quad e^* = c + d.
\end{equation}
This interval represents the unique non-degenerate stationary solution in the domain where $s<e$. 

\paragraph{Coverage computation.}
We can compute the coverage for this interval:
\begin{align*}
\int_{s^*}^{e^*} f(t)\,dt
\;&=\; \int_{c-d}^{c+d} \exp\Bigl(-\frac{(t-c)^2}{2\,w^2}\Bigr)\,dt \\
&= \sqrt{2}\,w \cdot \sqrt{\pi}\,\mathrm{erf}\bigl(\sqrt{-\ln(\lambda)}\bigr),
\end{align*}
where $\mathrm{erf}(\cdot)$ is the \emph{error function}.
Full derivations are included in the Appendix.
We denote this coverage as $C^*$. Meanwhile, the segment length is $e^* - s^*=2d = 2\,w\,\sqrt{-\,2\,\ln(\lambda)}$. Thus, using \eqref{eq:objective}, the objective value becomes:
\begin{align}
J(s^*, e^*) 
&= C^* - \lambda (e^* - s^*) \nonumber \\
&= \sqrt{2\pi}\,w \cdot \mathrm{erf}\!\left(\sqrt{-\ln(\lambda)}\right)
- \lambda \left(2w \sqrt{-2\ln(\lambda)}\right).
\label{eq:JStar}
\end{align}
For the regime where $0 \le \lambda < 1$, the coverage term is significantly larger than the segment length penalty term, resulting in a strictly positive objective value: $J(s^*, e^*) >0$.

\paragraph{Optimality for $0\le \lambda<1$.}
A non-degenerate interval 
\begin{equation}\label{eq:optimal-interval}
[s^*, e^*] = \left[c - d,\; c + d\right], \; \text{where } d = w\sqrt{-2\ln(\lambda)},
\end{equation}
is the unique stationary solution. 
This optimality is formally verified by observing that the objective function is concave due to the log-concavity of the Gaussian function.

\subsubsection{\texorpdfstring{$\bm{\lambda = 1}$}{lambda = 1}}
If $\lambda=1$, the condition in \eqref{eq:se-condition} implies:
\[
\exp\Bigl(-\frac{(s-c)^2}{2w^2}\Bigr) = 1 \quad\Longrightarrow\quad s=c,
\]
\[
\exp\Bigl(-\frac{(e-c)^2}{2w^2}\Bigr) = 1 \quad\Longrightarrow\quad e=c.
\]
Hence $s=e=c$, \ie the interval degenerates to a single point. In that case,
\[
\int_s^e f(t)\,dt = 0,\quad (e-s)=0,\quad \text{so }J(s,e)=0.
\]
For $\lambda = 1$, the bound in \eqref{eq:upperbound-objective} becomes $J(s,e) \le 0$. Equality holds only when $e=s=c$, yielding $J = 0$. Hence, the maximum value is zero, attained by the degenerate interval $[c, c]$.

\subsubsection{\texorpdfstring{$\bm{\lambda > 1}$}{lambda > 1}}
When $\lambda > 1$, the conditions $f(s) = \lambda$ and $f(e) = \lambda$ in \eqref{eq:se-condition} cannot be satisfied because $f(t) \le 1$ for all $t$. For any interval $[s, e]$ with positive length, the bound in \eqref{eq:upperbound-objective} yields
$
J(s, e) \le (1 - \lambda)(e - s),
$
which is strictly negative when $e - s > 0$ since $(1 - \lambda) < 0$.  
On the other hand, at $s = e = c$, we have $J = 0$.  
Therefore, the maximum value is $0$, attained only by the degenerate interval $[c, c]$.

{Overall Theorem}
We now revisit the objective in \eqref{eq:max-objective-eq} and combine all cases in \cref{sec:gbo-solution} to characterize the optimal segment:

\begin{theorem}[Optimal Segment Prediction]
\label{thm:optimal-seg}
Consider
\[
\max_{s,e}\; \Bigl[\;\int_{s}^{e} \bigl( f(t) \;-\; \lambda\bigr)\,dt\Bigr]
\]
where $f(t)$ is as in \eqref{eq:gaussian} and $\lambda\ge 0$ is the penalty weight.

\begin{enumerate}
    \item \textbf{If $0 \le \lambda < 1$:} A unique non-degenerate solution $[s^*,e^*]$ exists, centered at $c$, satisfying $f(s^*)=f(e^*)=\lambda$. In particular,
    $
    s^* = c - w\sqrt{-\,2\,\ln(\lambda)},\quad
    e^* = c + w\sqrt{-\,2\,\ln(\lambda)}.
    $
    \item \textbf{If $\lambda = 1$:} The optimal solution is the degenerate interval $[c,c]$, yielding $J_{\max}=0$. 
    \item \textbf{If $\lambda > 1$:} No positive-length interval can produce a nonnegative $J$. Hence $s=e=c$ (segment length $0$) is optimal with $J_{\max}=0$.
\end{enumerate}
\end{theorem}



\subsubsection{Interpretation of the Penalty Weight}
As shown in \eqref{eq:max-objective-eq}, introducing the constant baseline $\lambda$ inside the integrand reduces the net area of the shifted proposal as $\lambda$ increases. 
This motivates two complementary interpretations of the penalty weight:
\begin{itemize}
    \item \textbf{Penalty perspective.} $\lambda$ penalizes the net area of the shifted proposal, encouraging concise segment length. A larger $\lambda$ imposes stronger length regularization and typically yields a shorter interval.
    \item \textbf{Density perspective.}
    $\lambda$ serves as the horizontal cutoff level for the shifted proposal. Raising $\lambda$ lifts this cutoff line, so the part of the curve that remains above it becomes narrower but shifts toward the peak of the proposal, concentrating on its high-density region. 
    Thus, the chosen interval gets \emph{shorter} but \emph{denser}.
\end{itemize}

%

\subsection{GBO Extension to Gaussian Mixture Proposals}
\label{sec:mixture}
While previous methods such as CNM~\cite{zheng2022cnm} and CPL~\cite{zheng2022cpl} define each proposal using a single Gaussian, PPS~\cite{kim2024gaussian} introduces multiple Gaussians within a single proposal, resulting in a Gaussian mixture proposal.
To apply our GBO framework to such a Gaussian mixture proposal, we first identify the leftmost and rightmost Gaussians within the proposal. Let $\mathbf{c} = [c_1, \dots, c_{N}]$ and $\mathbf{w} = [w_1, \dots, w_{N}]$ be the predicted centers and widths of the $N$ Gaussians in a proposal.
We define the start and end points of each Gaussian as:
\begin{equation}
  s_n = c_{n} - \frac{1}{2}w_n, \;
  e_n = c_{n} + \frac{1}{2}w_{n}, \quad \text{for } n = 1, \dots, N.
\end{equation}
To determine the overall segment boundaries, we select the earliest start and the latest end among the $N$ components:
\begin{math}
  s_{\text{GMP}} = \min_{n} s_n, \quad e_{\text{GMP}} = \max_{n} e_n.
\end{math}
Using these, we define the effective center and width for the Gaussian mixture proposal as:
\begin{align}
  c_{\text{GMP}} = \frac{s_{\text{GMP}} + e_{\text{GMP}}}{2}, \quad w_{\text{GMP}} = e_{\text{GMP}} - s_{\text{GMP}}.
\end{align}
These center and width values are then substituted into ~\eqref{eq:optimal-interval} to predict the optimal temporal segments.

%% file: 4-experiment.tex

\subsection{Datasets}
\label{sec:datasets}
\subsubsection{ActivityNet Captions} The ActivityNet Captions dataset~\cite{krishna2017dense} consists of approximately 20,000 untrimmed YouTube videos paired with 100,000 corresponding sentence queries.
The dataset is divided into 10,009 videos for training, 4,917 for validation (further subdivided into $val_1$ and $val_2$ splits), and 5,044 for testing.
Following prior work~\cite{zhang2019cross}, we utilize the $val_2$ split as the test set since the ground truth annotations for the official test set are not publicly available.

\subsubsection{Charades-STA} The Charades-STA dataset~\cite{gao2017tall}, derived from the Charades dataset~\cite{sigurdsson2016hollywood}, is specifically designed for temporal video grounding tasks. It consists of 7,986 training videos and 1,863 testing videos depicting various indoor daily activities. 
In total, the dataset comprises 16,128 video-query pairs, with 12,408 used for training and 3,720 for testing.

\subsection{Evaluation Metrics}
\label{sec:evaluation-metrics}
We adopt the standard evaluation protocol proposed by Gao~\etal~\cite{gao2017tall} and utilize two evaluation metrics: `R@$n$,IoU=$m$' and `R@$n$,mIoU'. 
The first evaluation metric, `R@$n$,IoU=$m$', indicates the percentage of samples in which at least one of the top-$n$ predicted segments achieves a temporal Intersection over Union (tIoU) exceeding a threshold $m$. 
The second evaluation metric, `R@$n$,mIoU', computes the average of the highest tIoU values among the top-$n$ predicted segments for each sample.

\subsection{Implementation Details}
\label{sec:implementation-details}
We follow preprocessing settings from previous methods~\cite{chen2020look,lin2020weakly,zheng2022cnm} with 200 video chunks and 20-token queries.
For feature extraction, we utilize C3D~\cite{tran2015learning} features for ActivityNet Captions dataset and I3D~\cite{carreira2017quo} features for Charades-STA dataset.
We apply our Gaussian boundary optimization to three open-source Gaussian proposal-based methods: CNM~\cite{zheng2022cnm}\footnote{\url{https://github.com/minghangz/cnm}}, CPL~\cite{zheng2022cpl}\footnote{\url{https://github.com/minghangz/cpl}}, and PPS~\cite{kim2024gaussian}\footnote{\url{https://github.com/sunoh-kim/pps}}.
We use the penalty weight $\lambda$ to balance proposal coverage and segment compactness. 
Notably, $\lambda$ is the only tunable parameter in our framework, which represents a significant advantage compared to conventional deep learning approaches that typically require numerous hyperparameters.
We conduct a fine-grained search over $\lambda$ values ranging from 0.001 to 1.000 with a step size of 0.001.
This results in a total of 1,000 distinct $\lambda$ settings evaluated for each experiment.
We set the fixed penalty weight $\lambda$ for each GBO-enhanced method: $0.919$ for GBOCNM, $0.886$ for GBOCPL, and $0.883$ for GBOPPS on Charades-STA dataset, while using $0.938$ for GBOCNM, $0.909$ for GBOCPL, and $0.904$ for GBOPPS on ActivityNet Captions dataset.

\input{tables/tab-sota-comparisons}

\input{tables/tab-baselines}

\subsection{Comparison with State-of-the-Arts}
\label{sec:comparison-with-state-of-the-art-methods}

We compare our Gaussian Boundary Optimization (GBO) with existing state-of-the-art methods on both ActivityNet Captions and Charades-STA datasets, as shown in \cref{tab:sota-comparisons}.

\paragraph{Quantitative results.}
On both datasets, our GBO-enhanced methods consistently achieve highly competitive or superior performance compared to recent state-of-the-art approaches. On ActivityNet Captions, GBOPPS achieves 74.21\% in R@5 at IoU=0.5 and 45.51\% at IoU=0.7, outperforming state-of-the-art methods such as VGCI and DSRN. Similarly, on Charades-STA, GBOPPS achieves the best results across nearly all evaluation metrics, reaching up to 87.49\% in R@5 at IoU=0.5.
While ETC~\cite{li2025etc} reports strong results, it leverages additional annotations generated by a multi-modal large language model, making it not directly comparable to other methods under standard evaluation protocols.

\paragraph{Performance gains over baselines.}
To better understand the contribution of GBO, we quantify the performance improvements that GBO brings over its corresponding baselines (\ie CNM, CPL, and PPS) on both datasets, as shown in ~\cref{tab:baselines}.
Each GBO result is reported as a performance range (\eg 74.21\textendash75.42\%) to reflect two evaluation perspectives. 
The left value in each range represents performance achieved using a fixed penalty weight $\lambda$ that delivers strong and consistent results across all IoU thresholds.
In contrast, the right value indicates the maximum performance achieved by optimizing $\lambda$ specifically for each individual IoU threshold. 
Consequently, this right value can be interpreted as the upper bound of performance that our GBO framework can achieve.

GBO consistently enhances baseline performance across most evaluation metrics. For example, applying GBO to CPL yields substantial improvements up to +8.09\%p in R@5 at IoU=0.5 on ActivityNet Captions and up to +5.63\%p on Charades-STA. In the case of PPS, GBO achieves a remarkable performance gain of +11.25\%p in R@5 at IoU=0.7 on ActivityNet Captions. Notably, even when using a single fixed $\lambda$ value for all IoU thresholds, the improvements remain substantial, frequently exceeding +3\%p.
These results demonstrate the scalability and generalizability of our GBO framework across datasets and model architectures, confirming its effectiveness as a model-agnostic inference enhancement.

Notably, examining the performance ranges of our GBO methods reveals an important advantage.
The strong performance at both ends of each range demonstrates that our method maintains stability and effectiveness across various IoU thresholds when using a single well-chosen $\lambda$ value, while also offering the flexibility to be optimized for specific IoU thresholds when maximum performance is required.

\subsection{Analysis of GBO}
\label{sec:analysis-of-gbo}

\begin{figure*}[t!]
  \centering
  \includegraphics[width=0.75\linewidth]{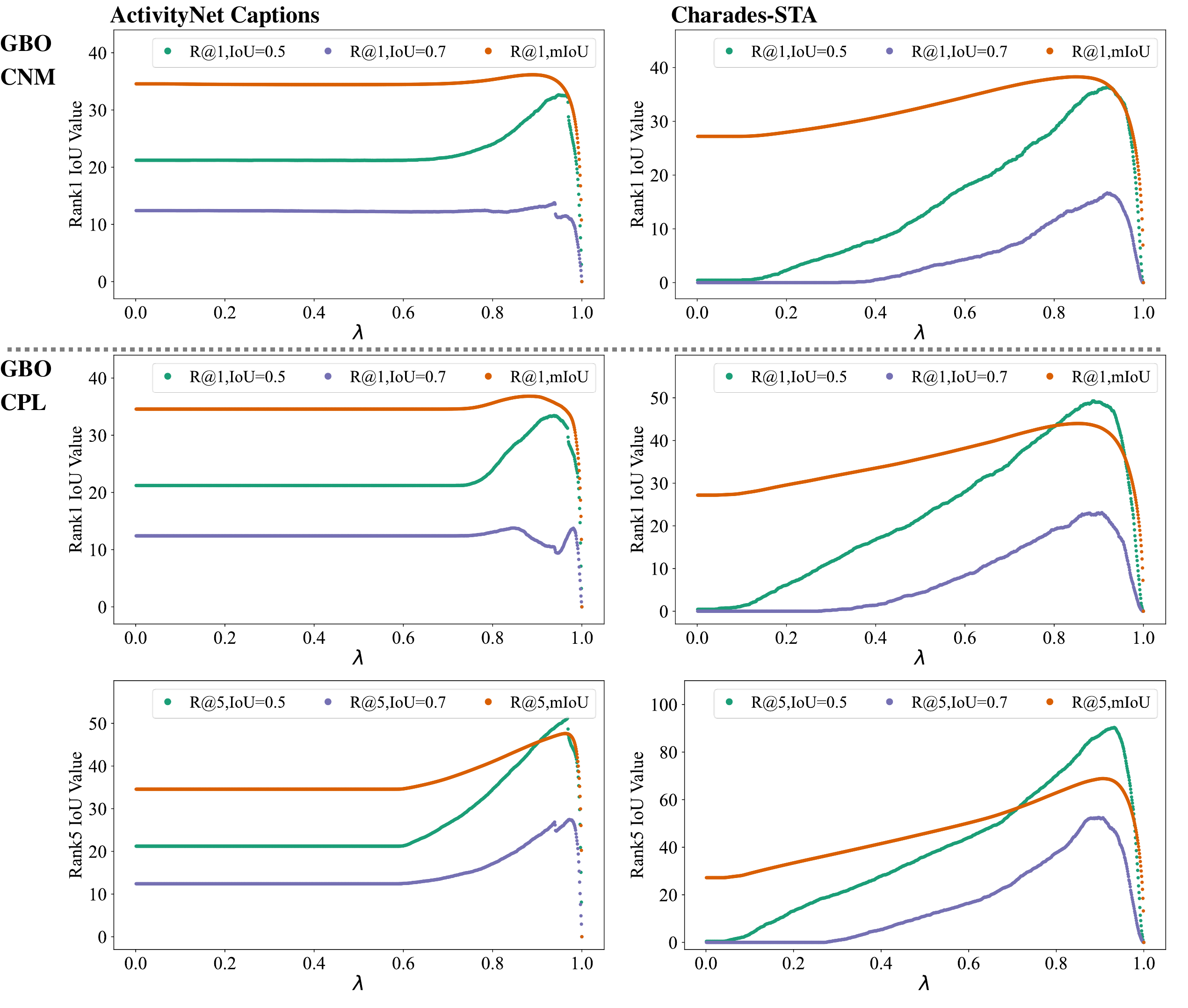}
  \caption{
    Performance comparison of segment predictions under different penalty weights $\lambda$. 
    The left and right columns correspond to ActivityNet Captions and Charades-STA, respectively. In each column, the top row shows results from GBOCNM, while the two bottom rows show results from GBOCPL at Rank@1 and Rank@5, respectively.
    Green, purple, and orange dots indicate performance at IoU=0.5, IoU=0.7, and mean IoU, respectively. 
  }
\label{fig:different-lambda}
\end{figure*}

\begin{figure*}[t!]
  \centering
  \includegraphics[width=0.75\linewidth]{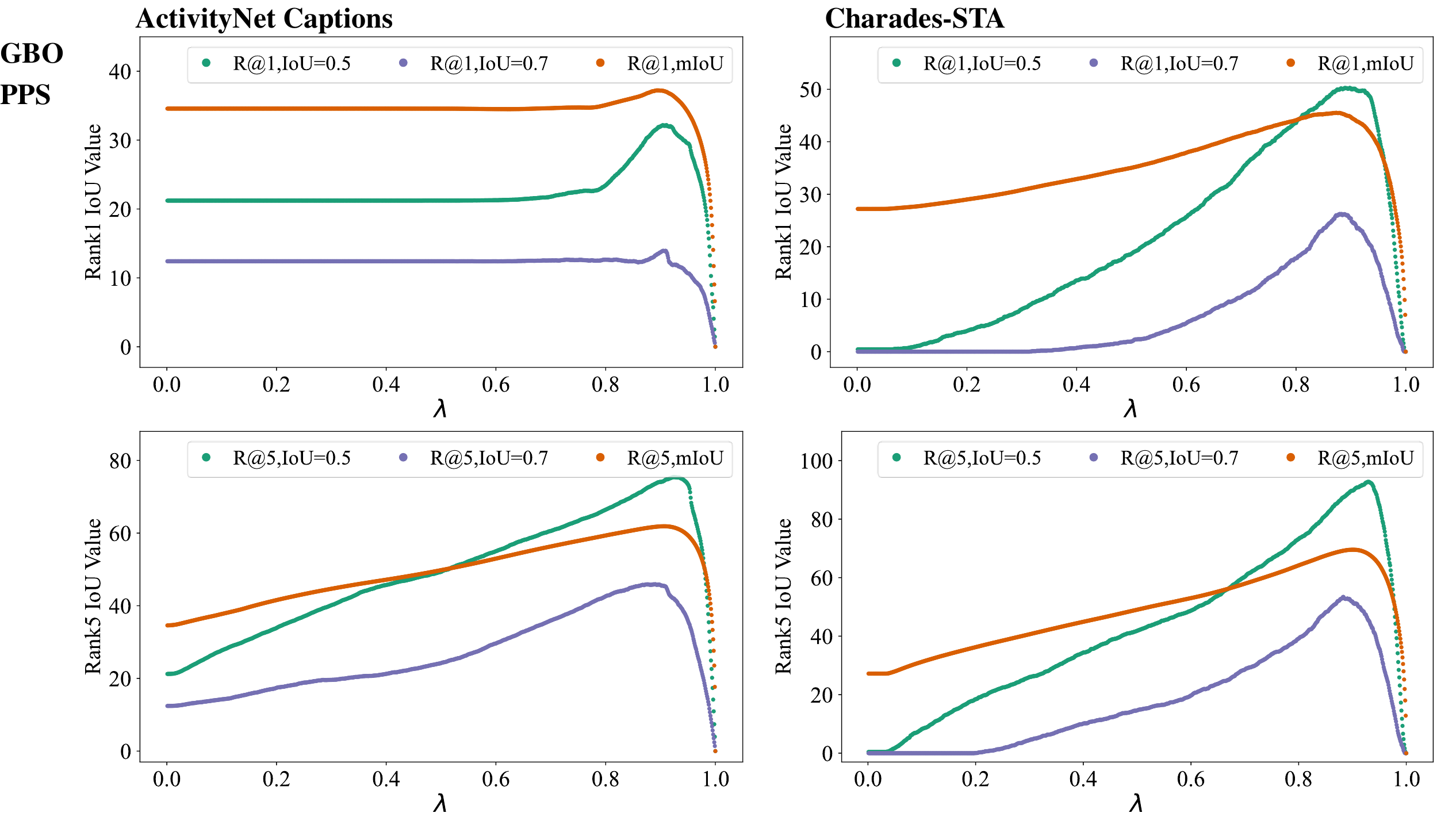}
  \caption{
    Performance comparison of segment predictions under different penalty weights $\lambda$. 
    The left and right columns correspond to ActivityNet Captions and Charades-STA, respectively. The two rows show results from GBOPPS at Rank@1 and Rank@5, respectively.
    Green, purple, and orange dots indicate performance at IoU=0.5, IoU=0.7, and mean IoU, respectively. 
  }
\label{fig:different-lambda-2}
\end{figure*}

\paragraph{Impact of $\lambda$.}
In \cref{fig:different-lambda} and \cref{fig:different-lambda-2}, we analyze the relationship between the penalty weights $\lambda$ and localization performance.
As $\lambda$ increases, our GBO method produces more compact segments that precisely capture the query-relevant content while effectively minimizing redundant context. 
However, when $\lambda$ approaches 1, we observe that the selected segment collapses into a degenerate interval, leading to a sharp drop in performance. 
This empirical finding highlights the importance of balancing coverage and compactness in temporal boundary prediction.

\paragraph{Existence of an optimal boundary region.}
Interestingly, in \cref{fig:different-lambda} and \cref{fig:different-lambda-2}, we observe that the peak performance values across different IoU thresholds and mean IoU consistently occur within a narrow range of $\lambda$ values. 
This consistent trend suggests the existence of an optimal boundary region that best matches the query-relevant content, regardless of the specific evaluation metric.
Our GBO formulation effectively identifies such regions by optimizing the trade-off between the coverage of the Gaussian proposal and the segment compactness.

\input{tables/tab-correlation}

\paragraph{Robustness across baselines.}
In \cref{fig:different-lambda} and \cref{fig:different-lambda-2}, we also observe consistent patterns in the performance curves across different GBO-enhanced variants~(\ie GBOCNM, GBOCPL, and GBOPPS), particularly on Charades-STA. 
To quantify this consistency, we compute the Pearson correlation coefficients between the performance curves of different GBO-enhanced variants, as shown in \cref{tab:correlation_vertical}. 
On Charades-STA, the correlations are relatively high (\eg 0.78 in R@1 at IoU=0.7 between GBO-PPS and GBO-CPL), indicating that the curves exhibit similar trends as $\lambda$ varies. 
Even on ActivityNet Captions, where the overall correlations are comparatively lower, the peak performance regions remain consistently aligned, supporting our claim that GBO reliably identifies optimal boundaries across baselines. 
This suggests that while GBO is generally robust, its sensitivity to $\lambda$ may depend on the quality and distribution of the underlying proposals.
Nevertheless, all variants consistently demonstrate performance improvements near a shared optimal region, reinforcing the generalizability of our GBO framework across diverse baselines.

\input{tables/tab-diverse-masks}
\input{tables/tab-top1-prediction}

\paragraph{Impact of top-1 prediction selection strategy.}
We analyze the effect of different top-1 prediction selection strategies, following several formulations proposed in \cite{kim2025enhancing}. 
Specifically, we evaluate four distinct strategies for scoring predictions: using only the reconstruction loss term (`Only Loss'), only the average IoU with other proposals (`Only IoU'), and combining both by weighting the IoU scores with loss values normalized by summation (`IoU+LossSum') and maximum (`IoU+LossMax').
The `Only IoU' strategy selects the proposal with the highest average IoU to other proposals, while `Only Loss' selects the proposal with the lowest query reconstruction loss~\cite{lin2020weakly}. 
The `IoU+LossSum' and `IoU+LossMax' strategies, proposed in \cite{kim2025enhancing}, apply semantic-aware weighting to the IoU scores using normalized loss values based on total sum and maximum value, respectively.
As shown in \cref{tab:top1-prediction}, we observe that the top-1 performance remains relatively stable across different selection strategies, indicating that our GBO framework exhibits robust performance regardless of the choice of prediction selection strategies.
Notably, `Only IoU' and `IoU+LossSum' strategies achieve nearly identical performance with a marginal difference below 0.01\%p. In contrast, `Only Loss' strategy underperforms by 2.50\%p in R@1 at IoU=0.7, highlighting the benefit of incorporating IoU-based information.

\input{tables/tab-inference-time}


\paragraph{Generalization beyond Gaussian proposals.}
Although our closed-form boundary solution is derived under a Gaussian assumption, \cref{tab:diverse-mask} shows that GBO consistently improves performance across diverse proposals.
On ActivityNet Captions, proposals such as Triangular and Epanechnikov achieve large gains of 5.57 and 5.66\%p at R@5,mIoU, while on Charades-STA, the Cauchy proposal yields strong improvements of 4.65 and 5.30 \%p at R@1 and R@5.
Notably, GBO shows no performance degradation under any setting.
These results confirm that, despite being analytically derived for Gaussian proposals, our method generalizes effectively to non-Gaussian representations as a proposal-agnostic inference module. Detailed proposal definitions are provided in the Appendix.
For each proposal, we train a CPL-based model by replacing the Gaussian proposal with the corresponding alternative on ActivityNet Captions and Charades-STA. 
We then directly apply our GBO during inference of these models to validate GBO's extensibility.

\begin{figure*}[t!]
  \centering
  \includegraphics[width=0.75\linewidth]{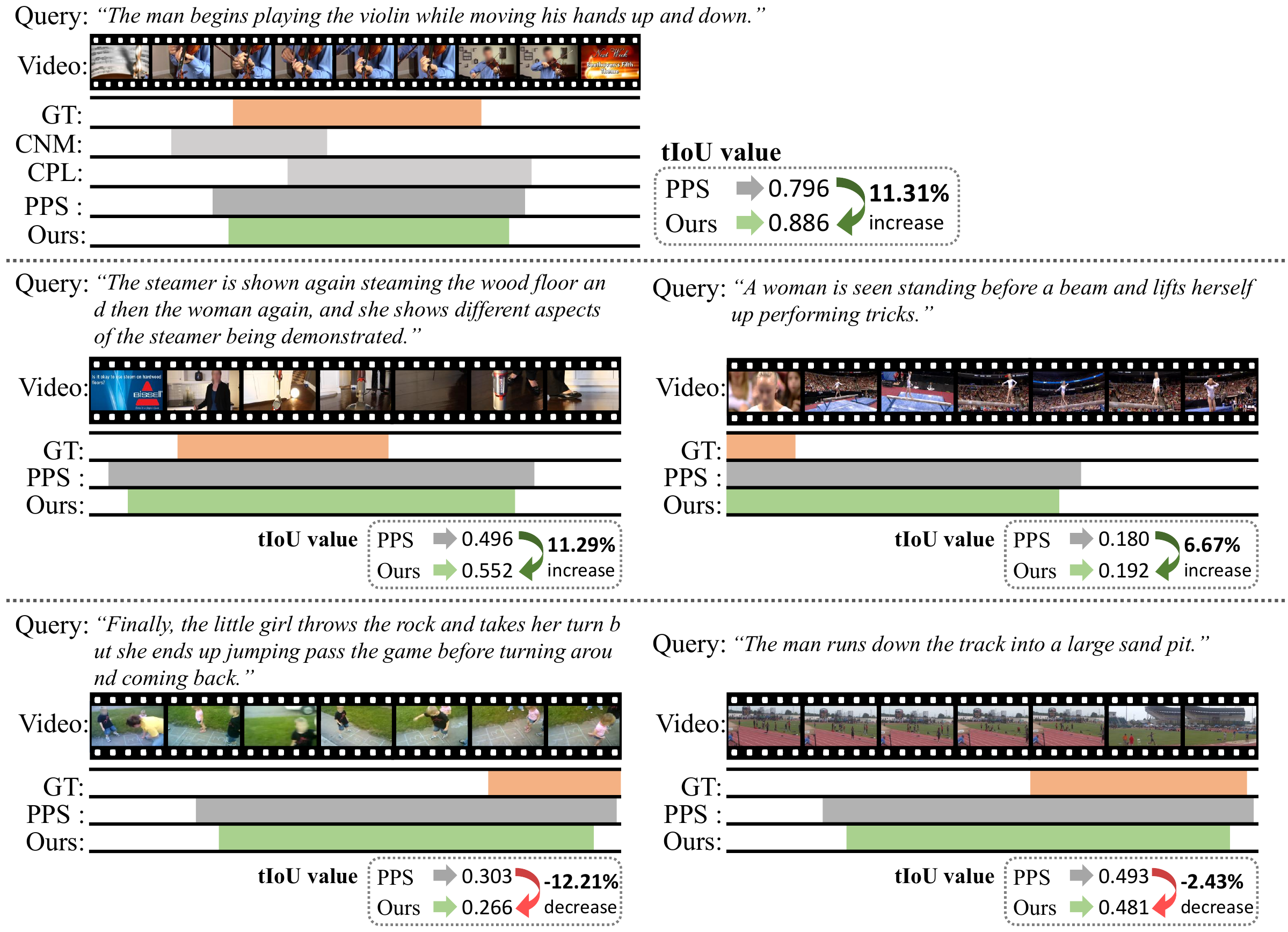}
  \caption{
    \textbf{Qualitative comparisons.}
    Each row shows the ground truth (GT), the baselines, and our GBOPPS prediction.
    Rows~1--2 are \emph{successful cases}, where symmetric compaction around the proposal removes redundant context and improves alignment, leading to higher tIoU.
    Row~3 is a \emph{failure case}: when the proposal center is misaligned and one PPS boundary coincides with the GT, our symmetric adjustment shifts both sides, displacing the well-aligned boundary and reducing overlap.
    Such cases are relatively rare, while in most scenarios ours yields more precise temporal localization.
  }
\label{fig:qualitative-result}
\end{figure*}

\paragraph{Efficiency Analysis.}
We evaluate inference efficiency with and without GBO across CNM, CPL, and PPS on a single NVIDIA GTX~1070, and report per-sample test time, the number of model parameters, and asymptotic per-sample complexity during inference in \cref{tab:inference-time-extended}.
GBO preserves model size and keeps runtime essentially unchanged, since GBO only applies a pre-computed closed-form update to each proposal for inference without introducing new parameters or iterative computation.
The measured per-sample overhead is below $0.3\%$ in all cases:
on Charades-STA, CNM from $0.828$ to $0.830$ ($\sim\!0.2\%$), CPL from $4.607$ to $4.611$ ($\sim\!0.1\%$), PPS from $7.796$ to $7.810$ ($\sim\!0.2\%$);
on ActivityNet Captions, CNM from $0.802$ to $0.803$ ($\sim\!0.1\%$), CPL from $5.544$ to $5.552$ ($\sim\!0.1\%$), PPS from $6.367$ to $6.384$ ($\sim\!0.3\%$).
These results highlight that GBO is a practical inference framework, delivering significant accuracy gains (\cref{tab:baselines}), with negligible runtime cost.

We further analyze asymptotic inference complexity. 
Let $P$ denote the number of proposals per sample.
\begin{itemize}
  \item \textbf{CNM: $\mathcal{O}(1)$.} CNM uses a single proposal and directly outputs the prediction, so inference cost is constant and independent of $P$.
  \item \textbf{CPL / PPS: $\mathcal{O}(P^{2}+P\log P)$.}
  Inference involves (i) sorting $P$ proposals by loss ($P\log P$), and (ii) pairwise IoU-based voting ($P^{2}$), after which top-$k$ proposals are selected using fixed $k$.
\end{itemize}
When combined with CNM, CPL, or PPS, GBO introduces only negligible asymptotic cost. 
Each proposal boundary is updated via a closed-form formula using a few arithmetic operations, which scales as $\mathcal{O}(P)$. 
In the case of CNM, where $P{=}1$, the update is constant-time, \ie $\mathcal{O}(1)$.
In the case of CPL and PPS, since the baseline pipelines already require $\mathcal{O}(P^{2}+P\log P)$, this additional $\mathcal{O}(P)$ term is negligible in practice, leading to similar runtimes as shown in \cref{tab:inference-time-extended}. 

When sweeping $L$ values of $\lambda$ for hyperparameter tuning, the cost scales as
$
\mathcal{O}\!\big(L(P^{2}+P\log P)\big),
$
which remains lightweight in practice because GBO depends only on a \emph{single} hyperparameter $\lambda$. 
Consequently, even with tuning, the additional computation is modest while the accuracy improvements are substantial.

\subsection{Qualitative Analysis}
\label{subsec:qualitative}

\cref{fig:qualitative-result} presents qualitative examples, comparing the localization results of the ground truth (GT), the baselines (CNM, CPL, and PPS), and our GBOPPS. The first two rows illustrate successful cases, while the last row depicts a failure case.

\paragraph{What changes in the successful cases.}
In the top two rows, given the query, the baseline PPS selects a relatively broad temporal segment that partially overlaps with the ground truth but contains redundant context. 
In contrast, our GBOPPS prediction aligns more precisely with the human-annotated moment, effectively capturing the core action described in the query while excluding irrelevant frames.
This qualitative trend is also reflected in the tIoU scores, \ie $0.796$ to $0.886$ ($+11.31\%$), $0.496$ to $0.552$ ($+11.29\%$), and $0.180$ to $0.192$ ($+6.67\%$).
These examples illustrate how our boundary optimization approach leads to more accurate temporal localization by achieving better alignment between predicted segments and actual query-relevant content.

\paragraph{Why the failure case happens.}
The failure case in the third row arises from the way GBOPPS adjusts segment boundaries. 
Our method always shifts both start and end points in a symmetric manner around the proposal center. 
This assumption is necessary to make the optimization tractable, and it usually removes redundant context and sharpens alignment, as demonstrated in the aforementioned successful cases.
However, in the third-row example, this adjustment becomes detrimental when only one boundary of the baseline already matches the ground truth while the other boundary does not, which typically happens because PPS misplaces the proposal center but coincidentally aligns one boundary.
In this case, a symmetric compaction toward the query-relevant center inevitably causes the well-aligned boundary to drift away, resulting in a mismatch. 
Nevertheless, such cases are rare, and in most other scenarios the symmetric adjustment improves grounding accuracy, as reported in \cref{tab:baselines}.
Moreover, this failure situation does not always degrade performance. 
In the second-row, right-hand example, although PPS is miscentered and one boundary happens to coincide with the GT, as in the failure cases.
In this case, the proposal extends beyond the opposite video boundary. Under symmetric compaction, the outward boundary is clipped at the video limit (0 seconds) while the inward boundary moves toward the query-relevant center. This effectively trims only redundant context and increases overlap ($+6.67\%$ tIoU).

\begin{table}[t!]
  \centering
  \caption{Comparison of average ground-truth (GT) moment lengths and average video lengths between Charades-STA and ActivityNet Captions.}
  \begin{tabular}{lcc}
    \toprule
    Dataset & Avg.\ GT length & Avg.\ video length \\
    \midrule
    Charades-STA & 8 sec & 30 sec \\
    ActivityNet Captions & 37 sec & 118 sec \\
    \bottomrule
  \end{tabular}
  \label{tab:gt-lengths}
\end{table}

\subsection{Discussion: Reason for some marginal performances on Charades-STA.}
\label{subsec:marginal-performance}
As shown in \cref{tab:baselines}, performances at R@1 metrics in Charades-STA are marginal or even drop.
The reasons are twofold. 
First, the top-1 prediction is often already close to the GT, leaving limited room for further improvement. 
Second, because ground-truth moments in Charades-STA are typically short (about 8 seconds), as shown in \cref{tab:gt-lengths}, the performance of top-1 prediction becomes highly sensitive to small boundary shifts. 
As a result, R@1 sometimes shows only marginal gains or slight decreases, particularly when the proposal center is misaligned while one boundary happens to coincide with the GT, which corresponds to the failure pattern in \cref{fig:qualitative-result}.

However, as shown in~\cref{tab:diverse-mask}, \emph{GBO consistently improves R@1,mIoU and R@5,mIoU}: +1.18\%p (R@1) and +4.58\%p (R@5) on ActivityNet Captions, and +0.29\%p (R@1) and +1.79\%p (R@5) on Charades-STA, respectively.
Although some performance at a fixed IoU threshold may appear marginal or lower, the overall ranking quality is enhanced, leading to higher top-1 and top-5 mIoU.
Moreover, R@5 benefits more strongly because GBOPPS refines every proposal, converting many near-misses into valid matches. 
This leads to clear gains in \cref{tab:baselines}: R@5 at IoU=0.5 improves from $84.71$ to $90.34$ (+5.63\%p) with GBOCPL and from $86.23$ to $92.75$ (+6.52\%p) with GBOPPS. 
Although improvements at IoU=0.7 are smaller on Charades-STA due to its shorter ground-truth segments (\cref{tab:gt-lengths}), GBOPPS still achieves the best performance across most evaluation metrics among state-of-the-art methods in \cref{tab:sota-comparisons}.


\subsection{Discussion: Potential Synergy with Decoupled Concept Prediction}\label{sec:dccp_discussion}
Decoupled consistent concept prediction (DCCP)~\cite{ma2022weakly} separates semantic concept prediction from boundary regression and enforces cross-video consistency, thereby reducing context bias in weakly supervised grounding. 
Our GBO focuses on refining temporal segments by optimizing proposal boundaries at inference. 
These two approaches are naturally complementary: DCCP provides cleaner, query-aligned semantic signals that can enhance the reliability of proposals, while GBO translates such improved proposal quality into sharper and more accurate boundary selection. 
Although we do not implement DCCP in this work, its potential integration highlights a promising avenue for future research.

\subsection{Discussion: Comparisons with Other Alternatives}
\noindent \textbf{Heuristic mapping}: Prior methods~\cite{kong2023dynamic,lv2023counterfactual,yoon2023scanet,bao2024omnipotent,kim2025enhancing,ma2023dual,lv2025variational,tang2025dual,li2025etc,zheng2022cnm,kim2024gaussian,zheng2022cpl} map center/width to $[s,e]$ by rule, underusing the proposal’s distribution and offering no optimality guarantee; \textbf{GT-IoU}: in weakly supervised settings, ground truth is unavailable at training/inference, so GT-IoU optimization is infeasible, and with oracle GT it trivially returns $[s,e]=[s_{\mathrm{GT}},e_{\mathrm{GT}}]$, \ie an upper bound; \textbf{Peer-IoU}: maximizing $J(s,e)=\tfrac{1}{K}\sum_{k=1}^{K}\operatorname{IoU}([s,e],P_k)$ captures only 1-D overlap, is non-smooth with no closed-form, and fails when only a single proposal exists (\eg CNM~\cite{zheng2022cnm}); by contrast, our method remains applicable with a single proposal by optimizing coverage relative to the proposal’s own distribution; \textbf{Reconstruction}: reconstruction loss~\cite{lin2020weakly,song2020weakly} reflects sentence reconstruction quality rather than proposal structure and reduces to a scalar. Related work (DIS~\cite{kim2025enhancing}) underperforms, as shown in \cref{tab:sota-comparisons}; \textbf{Our formulation} yields a smooth, log-concave objective with an interpretable threshold $f(s)=f(e)=\lambda$ and a closed-form optimum. It is model-agnostic, works with a single proposal, extends to mixtures and other proposals (\cref{tab:diverse-mask}), and consistently outperforms heuristic alternatives (\cref{tab:sota-comparisons}).

%% file: tables/tab-sota-comparisons.tex
\begin{table*}[t!]
  \centering
  \caption{Performance comparisons with state-of-the-art methods on ActivityNet Captions and Charades-STA.
  Bold and underlined values indicate the best and second-best results, respectively.
  } 
  {
  \begin{tabular}{c l | cc cc | cc cc}
    \toprule
    \multirow{3}{*}{Category} & \multirow{3}{*}{Method} & \multicolumn{4}{c}{ActivityNet Captions} & \multicolumn{4}{|c}{Charades-STA} \\
    & & \multicolumn{2}{c}{R@1} & \multicolumn{2}{c}{R@5} & \multicolumn{2}{|c}{R@1} & \multicolumn{2}{c}{R@5} \\ 
     & & IoU=0.5 & IoU=0.7 & IoU=0.5 & IoU=0.7 & IoU=0.5 & IoU=0.7 & IoU=0.5 & IoU=0.7 \\
    \midrule
    \multirow{7}{*}{\shortstack{Non-Gaussian\\Proposal}} & Random & 7.63 & - & 29.49 & - & 8.61 & 3.39 & 37.57 & 14.98 \\
    & CTF~\cite{chen2020look} & 23.60 & - & - & - & 27.30 & 12.90 & - & - \\
    & SCN~\cite{lin2020weakly} & 29.22 & - & 55.69 & - & 23.58 & 9.97 & 71.80 & 38.87 \\
    & MARN~\cite{song2020weakly} & 29.95 & - & 57.49 & - & 31.94 & 14.81 & 70.00 & 37.40 \\
    & CCL~\cite{zhang2020counterfactual} & 31.07 & - & 61.29 & - & 33.21 & 15.68 & 73.50 & 41.87 \\
    & VCA~\cite{wang2021visual} & 31.00 & - & 53.83 & - & 38.13 & 19.57 & 78.75 & 37.75 \\
    & CWSTG~\cite{Chen_Luo_Zhang_Ma_2022} & 29.52 & - & 66.61 & - & 31.02 & 16.53 & 77.53 & 41.91 \\
    \midrule
    \multirow{9}{*}{\shortstack{Gaussian\\Proposal\\(Closed)}} 
    & CPI~\cite{kong2023dynamic} & - &
    - & - & - & 50.47 &
    24.38 & 85.66 & 52.98 \\
    & CCR~\cite{lv2023counterfactual} & 30.39 & - & 56.50 & - & 50.79 & 23.75 & 84.48 & 52.44 \\
    & SCANet~\cite{yoon2023scanet} & 31.52 & - & 64.09 & - & 50.85 & 24.07 & 86.32 & \underline{53.28} \\
    & OmniD~\cite{bao2024omnipotent} & 31.60 & - & - & - & \textbf{52.31} & 24.35 & - & - \\
    & DIS~\cite{kim2025enhancing} & 30.30 & 12.38 & -  & - & 51.39 & 25.90 & - & - \\
    & DM2~\cite{ma2023dual} & 31.85 & - & - & - & 51.39 & 23.72 & - & - \\
    & VGCI~\cite{lv2025variational} & 31.55 & - & \underline{72.45}  & - & 50.71 & 25.84 & \underline{87.43} & 52.41 \\
    & DSRN~\cite{tang2025dual} & 30.29 & - & - & - & \underline{52.18} & 24.60 & - & - \\
    & ETC$^\dagger$~\cite{li2025etc} & \textcolor{gray}{37.01} & - & - & - & \textcolor{gray}{53.39} & \textcolor{gray}{25.84} & - & - \\
    \midrule
    \multirow{3}{*}{\shortstack{Gaussian\\Proposal\\(Open)}}& CNM~\cite{zheng2022cnm} & 28.71 & 12.82 & - & - & 35.18 & 14.95 & - & - \\
    & CPL~\cite{zheng2022cpl} & 31.37 & 11.83 & 43.13 & 22.95 & 49.24 & 22.39 & 84.71 & 52.37 \\
    & PPS~\cite{kim2024gaussian} & 31.25 & 12.34 & 71.32 & \underline{34.70} & 51.49 & \underline{26.16} & 86.23 & 53.01 \\
    \midrule
    \multirow{3}{*}{\shortstack{GBO\\(Ours)}}& GBOCNM
    & 32.14 & \underline{13.77} & - & - & 36.38 & 16.69 & -  & - \\
    & GBOCPL
    & \textbf{32.71} & 11.15 & 46.29 & 24.04 & 49.27 & 22.64 & 85.78 & 52.06 \\
    & GBOPPS
    & \underline{32.18} & \textbf{13.92} & \textbf{74.21} & \textbf{45.51} & 50.01 & \textbf{26.25} & \textbf{87.49} & \textbf{53.45} \\
    \bottomrule  
  \end{tabular}}
  \\[0.5ex]
    \begin{minipage}{0.8\linewidth}
    \raggedright
    {\small The method marked with $^\dagger$ utilizes additional annotations generated by a multi-modal large language model.}
    \end{minipage}
  \label{tab:sota-comparisons}
\end{table*}

%% file: tables/tab-baselines.tex
\begin{table*}[t!]
  \centering
  \caption{Performance improvements of Gaussian proposal-based methods via Gaussian boundary optimization on ActivityNet Captions and Charades-STA.
  Each GBO result is shown as a range, where the left value denotes performance with a single, fixed $\lambda$, and the right value indicates the best score across all $\lambda$ values.} 
  \resizebox{\linewidth}{!}
  {
  \begin{tabular}{l | cc cc | cc cc}
    \toprule
    \multirow{3}{*}{Method} & \multicolumn{4}{c}{ActivityNet Captions} & \multicolumn{4}{|c}{Charades-STA} \\
    & \multicolumn{2}{c}{R@1} & \multicolumn{2}{c}{R@5} & \multicolumn{2}{|c}{R@1} & \multicolumn{2}{c}{R@5} \\ 
    & IoU=0.5 & IoU=0.7 & IoU=0.5 & IoU=0.7 & IoU=0.5 & IoU=0.7 & IoU=0.5 & IoU=0.7 \\
    \midrule
    CNM~\cite{zheng2022cnm} & 28.71 & 12.82 & - & - & 35.18 & 14.95 & - & - \\
    GBOCNM
    & 32.14\textendash32.65 & {13.77}\textendash13.77 & - & - & 36.38\textendash36.42 & 16.69\textendash16.69 & -  & - \\
    \small{Improvement} & {\small\textcolor{teal}{+3.43}\textendash\textcolor{teal}{+3.94}} & {\small\textcolor{teal}{+0.95}\textendash\textcolor{teal}{+0.95}} & {\small-} & {\small-} & {\small\textcolor{teal}{+1.20}\textendash\textcolor{teal}{+1.24}} & {\small\textcolor{teal}{+1.74}\textendash\textcolor{teal}{+1.74}} & {\small-} & {\small-} \\
    \midrule
    CPL~\cite{zheng2022cpl} & 31.37 & 11.83 & 43.13 & 22.95 & 49.24 & 22.39 & 84.71 & 52.37 \\
    GBOCPL
    & {32.71}\textendash33.46 & 11.15\textendash13.77 & 46.29\textendash51.22 & 24.04\textendash27.47 & 49.27\textendash49.30 & 22.64\textendash23.12 & 85.78\textendash90.34 & 52.06\textendash52.53 \\
    \small{Improvement} & {\small\textcolor{teal}{+1.34}\textendash\textcolor{teal}{+2.09}} & {\small\textcolor{brown}{-0.68}\textendash\textcolor{teal}{+1.94}} & {\small\textcolor{teal}{+3.16}\textendash\textcolor{blue}{+8.09}} & {\small\textcolor{teal}{+1.09}\textendash\textcolor{teal}{+4.52}} & {\small\textcolor{teal}{+0.03}\textendash\textcolor{teal}{+0.06}} & {\small\textcolor{teal}{+0.25}\textendash\textcolor{teal}{+0.73}} & {\small\textcolor{teal}{+1.07}\textendash\textcolor{blue}{+5.63}} & {\small\textcolor{brown}{-0.31}\textendash\textcolor{teal}{+0.16}} \\
    \midrule
    PPS~\cite{kim2024gaussian} & 31.25 & 12.34 & 71.32 & {34.70} & 51.49 & {26.16} & 86.23 & 53.01 \\
    GBOPPS
    & {32.18}\textendash32.18 & {13.92}\textendash13.95 & {74.21}\textendash75.42 & {45.51}\textendash45.95 & 50.01\textendash50.28 & {26.25}\textendash26.25 & {87.49}\textendash92.75 & {53.45}\textendash53.45 \\
    \small{Improvement} & {\small\textcolor{teal}{+0.93}\textendash\textcolor{teal}{+0.93}} & {\small\textcolor{teal}{+1.58}\textendash\textcolor{teal}{+1.61}} & {\small\textcolor{teal}{+2.89}\textendash\textcolor{teal}{+4.10}} & {\small\textcolor{blue}{+10.81}\textendash\textcolor{blue}{+11.25}} & {\small\textcolor{brown}{-1.48}\textendash\textcolor{brown}{-1.21}} & {\small\textcolor{teal}{+0.09}\textendash\textcolor{teal}{+0.09}} & {\small\textcolor{teal}{+1.26}\textendash\textcolor{blue}{+6.52}} & {\small\textcolor{teal}{+0.44}\textendash\textcolor{teal}{+0.44}} \\
    \bottomrule  
  \end{tabular}}
  \label{tab:baselines}
\end{table*}

%% file: tables/tab-correlation.tex
\begin{table*}[t!]
  \centering
  \caption{Pearson correlation coefficients between performance curves of GBO-enhanced method pairs on ActivityNet Captions and Charades-STA. Higher values indicate stronger similarity in how different methods respond to varying $\lambda$ values.}
  \begin{tabular}{l | ccc | ccc}
    \toprule
    \multirow{2}{*}{Metric} 
    & \multicolumn{3}{c|}{ActivityNet Captions} 
    & \multicolumn{3}{c}{Charades-STA} \\
    & GBO PPS$\leftrightarrow$CPL & GBO PPS$\leftrightarrow$CNM & GBO CPL$\leftrightarrow$CNM
    & GBO PPS$\leftrightarrow$CPL & GBO PPS$\leftrightarrow$CNM & GBO CPL$\leftrightarrow$CNM \\
    \midrule
    R@1,IoU=0.5   & 0.3740 & 0.1063 & 0.2718 & 0.7206 & 0.4235 & 0.5061 \\
    R@1,IoU=0.7   & 0.2771 & 0.0735 & 0.1009 & 0.7836 & 0.3487 & 0.4530 \\
    R@1,mIoU  & 0.2716 & 0.1783 & 0.2781 & 0.6270 & 0.4803 & 0.5232 \\
    R@5,IoU=0.5   & 0.3257 & -      & -      & 0.6448 & -      & -      \\
    R@5,IoU=0.7   & 0.2040 & -      & -      & 0.7008 & -      & -      \\
    R@5,mIoU  & 0.3398 & -      & -      & 0.6572 & -      & -      \\
    \bottomrule
  \end{tabular}
  \label{tab:correlation_vertical}
\end{table*}

%% file: tables/tab-diverse-masks.tex
\begin{table*}[t]
\centering
\caption{Performance comparisons of various proposals on ActivityNet Captions and Charades-STA with/without GBO. For \emph{w/ GBO}, the value in parentheses is the absolute gain over \emph{w/o GBO} in percentage points.}
\label{tab:diverse-mask}
\resizebox{0.95\textwidth}{!}{
\setlength{\tabcolsep}{5pt}
\begin{tabular}{lcccccccc}
\toprule
\multirow{3}{*}{Proposal} & \multicolumn{4}{c}{ActivityNet Captions} & \multicolumn{4}{c}{Charades-STA} \\
 & \multicolumn{2}{c}{R@1,mIoU} & \multicolumn{2}{c}{R@5,mIoU} & \multicolumn{2}{c}{R@1,mIoU} & \multicolumn{2}{c}{R@5,mIoU} \\
 & w/o GBO & w/ GBO & w/o GBO & w/ GBO & w/o GBO & w/ GBO & w/o GBO & w/ GBO \\
\midrule
Gauss    & 35.38 & 36.56 (\textcolor{teal}{+1.18\%p}) & 41.33 & 45.91 (\textcolor{teal}{+4.58\%p}) & 44.70 & 44.70 (\textcolor{teal}{+0.00\%p}) & 67.67 & 68.46 (\textcolor{teal}{+0.79\%p}) \\
Laplace  & 36.52 & 36.58 (\textcolor{teal}{+0.07\%p}) & 43.12 & 45.87 (\textcolor{teal}{+2.75\%p}) & 42.44 & 43.17 (\textcolor{teal}{+0.72\%p}) & 68.34 & 68.39 (\textcolor{teal}{+0.05\%p}) \\
Cauchy   & 35.56 & 35.87 (\textcolor{teal}{+0.31\%p}) & 42.90 & 44.65 (\textcolor{teal}{+1.75\%p}) & 34.98 & 39.63 (\textcolor{teal}{+4.65\%p}) & 61.00 & 66.30 (\textcolor{teal}{+5.30\%p}) \\
Logistic & 36.52 & 36.53 (\textcolor{teal}{+0.01\%p}) & 43.02 & 44.19 (\textcolor{teal}{+1.18\%p}) & 42.25 & 42.80 (\textcolor{teal}{+0.55\%p}) & 68.56 & 68.56 (\textcolor{teal}{+0.00\%p}) \\
Triangular      & 34.69 & 34.71 (\textcolor{teal}{+0.02\%p}) & 44.12 & 49.69 (\textcolor{teal}{+5.57\%p}) & 43.34 & 43.59 (\textcolor{teal}{+0.25\%p}) & 65.09 & 68.64 (\textcolor{teal}{+3.55\%p}) \\
Epanechnikov      & 34.96 & 34.96 (\textcolor{teal}{+0.00\%p}) & 44.11 & 49.77 (\textcolor{teal}{+5.66\%p}) & 43.19 & 43.40 (\textcolor{teal}{+0.21\%p}) & 66.96 & 68.65 (\textcolor{teal}{+1.70\%p}) \\
Student's-$t$ & 36.79 & 36.79 (\textcolor{teal}{+0.00\%p}) & 42.93 & 44.85 (\textcolor{teal}{+1.92\%p}) & 40.08 & 42.31 (\textcolor{teal}{+2.24\%p}) & 66.83 & 67.38 (\textcolor{teal}{+0.55\%p}) \\
Rational Quadratic & 36.05 & 36.52 (\textcolor{teal}{+0.47\%p}) & 42.85 & 47.17 (\textcolor{teal}{+4.32\%p}) & 41.81 & 42.27 (\textcolor{teal}{+0.45\%p}) & 68.01 & 68.24 (\textcolor{teal}{+0.24\%p}) \\
\bottomrule
\end{tabular}}
\end{table*}

%% file: tables/tab-top1-prediction.tex
\begin{table}[t!]
  \centering
  \caption{Ablation study on different top-1 prediction selection strategies for GBOPPS on ActivityNet Captions.}
  \begin{tabular}{l | cc  cc}
    \toprule
    \multirow{2}{*}{Strategy} & \multicolumn{2}{c}{R@1} & \multicolumn{2}{c}{R@5}\\ 
    & IoU=0.5 & IoU=0.7 & IoU=0.5 & IoU=0.7 \\
    \midrule
    Only Loss & 27.40 & 11.42 & 74.21 & 45.51 \\
    Only IoU & 32.18 & 13.92 & 74.21 & 45.51  \\
    IoU+LossMax & 29.62 & 11.27 & 74.21 & 45.51 \\
    IoU+LossSum & 32.18 & 13.92 & 74.21 & 45.51 \\
    \bottomrule  
  \end{tabular}
  \label{tab:top1-prediction}
\end{table}

%% file: tables/tab-inference-time.tex

\begin{table*}[t]
  \centering
  \caption{Inference characteristics with and without Gaussian Boundary Optimization (GBO). For each method, we report per-sample inference time, the number of model parameters, and asymptotic per-sample complexity on Charades-STA and ActivityNet Captions. Here, $P$ is the number of proposals. GBO preserves model size and maintains essentially identical runtime.}
  \label{tab:inference-time-extended}
  \renewcommand{\arraystretch}{1.1}
  \resizebox{0.95\textwidth}{!}{%
  \setlength{\tabcolsep}{6pt}
  \begin{tabular}{l |c c l | c c l}
    \toprule
    \multirow{2}{*}{Method} & \multicolumn{3}{c|}{Charades-STA} & \multicolumn{3}{c}{ActivityNet Captions} \\
     & Per-sample Time & \# Parameters & Inference Complexity & Per-sample Time & \# Parameters & Inference Complexity \\
    \midrule
    CNM      & 0.828\,ms & 5.37\,M  & $\mathcal{O}(1)$ & 0.802\,ms & 7.01\,M  & $\mathcal{O}(1)$ \\
    GBOCNM   & 0.830\,ms & 5.37\,M  & $\mathcal{O}(1)$ & 0.803\,ms & 7.01\,M  & $\mathcal{O}(1)$ \\
    \midrule
    CPL      & 4.607\,ms & 5.38\,M  & $\mathcal{O}(P^{2} + P\log P),\ P{=}8$ & 5.544\,ms & 7.01\,M  & $\mathcal{O}(P^{2} + P\log P),\ P{=}8$ \\
    GBOCPL   & 4.611\,ms & 5.38\,M  & $\mathcal{O}(P^{2} + P\log P),\ P{=}8$ & 5.552\,ms & 7.01\,M  & $\mathcal{O}(P^{2} + P\log P),\ P{=}8$ \\
    \midrule
    PPS      & 7.796\,ms & 15.02\,M & $\mathcal{O}(P^{2} + P\log P),\ P{=}7$ & 6.367\,ms & 16.64\,M & $\mathcal{O}(P^{2} + P\log P),\ P{=}5$ \\
    GBOPPS   & 7.810\,ms & 15.02\,M & $\mathcal{O}(P^{2} + P\log P),\ P{=}7$ & 6.384\,ms & 16.64\,M & $\mathcal{O}(P^{2} + P\log P),\ P{=}5$ \\
    \bottomrule
  \end{tabular}}
\end{table*}

%% file: 5-conclusion.tex
We introduced Gaussian Boundary Optimization (GBO), a principled inference framework for weakly supervised temporal video grounding. Unlike existing Gaussian-based methods that rely on heuristic inference strategies, GBO formulates segment prediction as an optimization problem that balances proposal coverage and segment compactness. We provide a closed-form solution to this optimization problem and rigorously characterize the optimal segment structure under different regimes of the penalty weight~$\lambda$. This yields clear interpretability: GBO predicts a non-degenerate interval for $0 \le \lambda < 1$, and a degenerate point for $\lambda \ge 1$.
Importantly, GBO requires only a single interpretable hyperparameter $\lambda$ that enables fine-grained control over segment boundaries without necessitating complex tuning processes.
Our framework demonstrates model-agnostic properties, seamlessly integrating with both single-Gaussian and Gaussian mixture proposals and showing consistent improvements across diverse architectures without requiring additional training or inference overhead.
While our formulation is applicable to various proposal types, it assumes symmetric boundary adjustment, which may limit its effectiveness for highly asymmetric temporal events. Future work will explore asymmetric boundary optimization schemes to further enhance generalization across diverse temporal structures.

%% file: 6-acknowledgements.tex
The present research was supported by the research fund of Dankook University in 2025.

%% file: 7-appendix.tex
{\appendices
\section*{Proof of the Stationary Condition}

\paragraph{Derivative with respect to $e$.}
\begin{align*}
\frac{\partial}{\partial e}\bigl[\int_s^e f(t)\,dt \bigr] 
= f(e), \;
\frac{\partial}{\partial e}\bigl[-\lambda (e-s)\bigr] 
= -\lambda.
\end{align*}
Hence,
\[
\frac{\partial J}{\partial e}
\;=\; f(e) \;-\; \lambda.
\]
Setting $\frac{\partial J}{\partial e} = 0$ yields:
\begin{math}
f(e) \;=\; \lambda.
\end{math}

\paragraph{Derivative with respect to $s$.}

A similar calculation for the derivative with respect to \( s \) gives:
\begin{math}
f(s) \;=\; \lambda.
\end{math}

These results together imply that \emph{any} stationary point within the domain $s<e$ must satisfy
\begin{math}
f(s) \;=\; f(e) \;=\; \lambda.
\end{math}

\section*{Derivation of Coverage Expression}

We can compute the coverage for the interval given by \eqref{eq:optimal-boundary}:
\[
\int_{s^*}^{e^*} f(t)\,dt
\;=\; \int_{c-d}^{c+d} \exp\Bigl(-\frac{(t-c)^2}{2\,w^2}\Bigr)\,dt.
\]

Make the change of variable $u = \frac{t-c}{\sqrt{2}\,w}$. Then $du = dt/(\sqrt{2}\,w)$, and when $t=c\pm d$, $u=\pm \tfrac{d}{\sqrt{2}w} = \pm \sqrt{-\ln(\lambda)}$. Thus:
\begin{align*}
\int_{c-d}^{c+d} 
\exp\!\Bigl(-\frac{(t-c)^2}{2\,w^2}\Bigr)\,dt
&= \sqrt{2}\,w \int_{-\,\sqrt{-\ln(\lambda)}}^{+\,\sqrt{-\ln(\lambda)}} \exp(-u^2)\,du.
\end{align*}
Recall the \emph{error function}:
\[
\mathrm{erf}(z)
= \frac{2}{\sqrt{\pi}}\int_{0}^{z} \exp(-t^2)\,dt,
\]
and the related identity:
\[
\int_{-a}^{+a} \exp(-u^2)\,du
= \sqrt{\pi}\,\mathrm{erf}(a).
\]
Applying this identity to the integral, we obtain:
\[
\int_{-\,\sqrt{-\ln(\lambda)}}^{+\,\sqrt{-\ln(\lambda)}} \exp(-u^2)\,du
= \sqrt{\pi}\,\mathrm{erf}\bigl(\sqrt{-\ln(\lambda)}\bigr).
\]
Substituting this result back, the coverage becomes:
\begin{align*}
\int_{c-d}^{c+d} f(t)\,dt
&= \sqrt{2}\,w \cdot \sqrt{\pi}\,\mathrm{erf}\bigl(\sqrt{-\ln(\lambda)}\bigr).
\end{align*}

\section*{Proposal Definitions}
\label{sec:masks}

We employ a family of learnable proposals \(M(t)\), each centered at \(c\) with width \(w>0\), where the normalized coordinate is \(t \in [0,1]\).
All proposals satisfy \(M(c)=1\).
For implementation, we apply an effective scale \(w/\sigma\) with a positive constant \(\sigma>0\).

\paragraph{Gaussian proposal.} 
This proposal exhibits the standard bell-shaped form with exponentially decaying tails. 
We adopted this proposal as the default choice in all theoretical analyses and experimental evaluations.
\[
M_{\mathrm{Gauss}}(t)
=\exp\!\Bigl(-\frac{(t-c)^2}{2\,(w/\sigma_{\mathrm{gauss}})^2}\Bigr).
\]

\paragraph{Laplace proposal.}
This proposal has a sharper peak than the Gaussian proposal and a steeper slope at the center. 
The tails decay exponentially.
\[
M_{\mathrm{Laplace}}(t)
=\exp\!\Bigl(-\frac{|t-c|}{\,w/\sigma_{\mathrm{laplace}}}\Bigr).
\]
    
\paragraph{Cauchy proposal.}
This proposal is characterized by very heavy tails and decreases slowly as the distance from the center increases.
\[
M_{\mathrm{Cauchy}}(t)
=\frac{1}{1+\Bigl(\tfrac{t-c}{\,w/\sigma_{\mathrm{cauchy}}}\Bigr)^{\!2}}.
\]

\paragraph{Triangular proposal.}
This proposal follows a piecewise linear pyramid-like profile with compact support 
\(\,|t-c|\le w/\sigma_{\mathrm{tri}}\).
\[
M_{\mathrm{Tri}}(t)
=\max\!\Bigl(1-\Bigl|\tfrac{t-c}{\,w/\sigma_{\mathrm{tri}}}\Bigr|,\,0\Bigr).
\]

\paragraph{Epanechnikov proposal.}
This proposal exhibits a parabolic profile with compact support 
\(\,|t-c|\le w/\sigma_{\mathrm{epa}}\), 
and has a flatter interior compared to the triangular proposal.
\[
M_{\mathrm{Epa}}(t)
=\max\!\Bigl(1-\Bigl(\tfrac{t-c}{\,w/\sigma_{\mathrm{epa}}}\Bigr)^{\!2},\,0\Bigr).
\]

\paragraph{Logistic proposal.}
This proposal resembles the Gaussian proposal around the center but has slightly broader shoulders.
\[
M_{\mathrm{Log}}(t)
=4\,\sigma(z)\bigl(1-\sigma(z)\bigr),
\;\;
z=\tfrac{t-c}{\,w/\sigma_{\mathrm{log}}},\;
\sigma(z)=\tfrac{1}{1+e^{-z}}.
\]

\paragraph{Student's-\(t\) proposal.}
This proposal behaves similarly to a Gaussian near the peak, but the heavier tails are controlled by the degrees of freedom parameter \(\nu>2\).
\[
M_{\mathrm{St}}(t)
=\Biggl(1+\frac{(t-c)^2}{\,\nu\,(w/\sigma_{\mathrm{t}})^2}\Biggr)^{-\frac{\nu+1}{2}},
\;\;
\sigma_{\mathrm{t}}=4\sqrt{\frac{\nu}{\nu-2}}.
\]

\paragraph{Rational quadratic (RQ) proposal.}
This proposal smoothly interpolates between Gaussian-like behavior in the limit \(\alpha\to\infty\) and heavy-tailed behavior for smaller values of \(\alpha>0\).
\[
M_{\mathrm{RQ}}(t)
=\Biggl(1+\frac{(t-c)^2}{2\alpha(w/\sigma_{\mathrm{rq}})^2}\Biggr)^{-\alpha},\,
\sigma_{\mathrm{rq}}=8\sqrt{2\alpha\bigl(2^{1/\alpha}-1\bigr)}.
\]

\paragraph{Proposal parameter.}
For the proposal generator, the scale parameter $\sigma$ is used to map the predicted width \(w\) to the effective scale \(w/\sigma\) in each proposal.
Following~\cite{zheng2022cpl}, we set the Gaussian scale to 
\(\sigma_{\mathrm{gauss}}=9\). 
All other constants are defined relative to this baseline value:
$
\sigma_{\mathrm{laplace}}=\sigma_{\mathrm{cauchy}}=\sigma_{\mathrm{gauss}},\quad
\sigma_{\mathrm{tri}}=\tfrac{1}{2}\sigma_{\mathrm{gauss}},\quad
\sigma_{\mathrm{epa}}=\tfrac{1}{2}\sigma_{\mathrm{gauss}},\quad
\sigma_{\mathrm{log}}=\tfrac{3}{2}\sigma_{\mathrm{gauss}},\quad
\nu=\tfrac{1}{4}\sigma_{\mathrm{gauss}},\quad
\alpha=\tfrac{1}{2}\sigma_{\mathrm{gauss}}.
$
All constants above are fixed during training and inference.

\paragraph{Optimization of other proposal types.}
While the current formulation of GBO is derived under the Gaussian assumption, similar optimization principles can be extended to other proposal types. Although deriving closed-form solutions for these proposals is challenging, similar formulations are achieved by utilizing their inherent symmetry around the center, as in the Gaussian case. A more rigorous mathematical treatment of these extensions is left as future work.